\documentclass[letterpaper]{article} % DO NOT CHANGE THIS
\usepackage[]{aaai2027}  % DO NOT CHANGE THIS
\usepackage[hyphens]{url}  % DO NOT CHANGE THIS
\usepackage{graphicx} % DO NOT CHANGE THIS
\usepackage{natbib}  % DO NOT CHANGE THIS AND DO NOT ADD ANY OPTIONS TO IT
\usepackage{caption} % DO NOT CHANGE THIS AND DO NOT ADD ANY OPTIONS TO IT
\usepackage{algorithm}
\usepackage{algorithmic}
\usepackage{multirow}
\usepackage{hhline}
\usepackage{xcolor}
\usepackage{colortbl}
\usepackage{amsmath}
\usepackage{amssymb}
\usepackage{cancel}
\usepackage{newfloat}
\usepackage{listings}
\DeclareCaptionStyle{ruled}{labelfont=normalfont,labelsep=colon,strut=off} % DO NOT CHANGE THIS
\floatstyle{ruled}
\newfloat{listing}{tb}{lst}{}
\floatname{listing}{Listing}

\usepackage{booktabs}

\title{LiveVVT: High-Fidelity Video Virtual Try-On in Real Time}
\author{
    Yushe Cao\textsuperscript{\rm 1},
    Shikun Feng\corresponding\textsuperscript{\rm 2},
    Ruxiang Duan\textsuperscript{\rm 3},
    Liyong Wang\textsuperscript{\rm 4}, \\
    Dianxi Shi\corresponding\textsuperscript{\rm 1},
    Chun Yu\textsuperscript{\rm 1},
    Junliang Xing\corresponding\textsuperscript{\rm 1},
}
\affiliations{
    \textsuperscript{\rm 1}Tsinghua University
    \textsuperscript{\rm 2}Zhongguancun Academy \\
    \textsuperscript{\rm 3}South China University of Technology
    \textsuperscript{\rm 4}Beijing Jiaotong University \\
    cao-ys23@mails.tsinghua.edu.cn,kunayumi@163.com
}

\begin{document}

\maketitle

% !TeX root = ../main.tex

\begin{abstract}
Diffusion-based Video Virtual Try-On (VVT) achieves high visual fidelity through bidirectional spatio-temporal modeling, but complete-clip dependence incurs prohibitive latency and computational overhead in practical continuous deployment. Naively enforcing causality disrupts pretrained bidirectional priors and substantially degrades synthesis quality. We introduce LiveVVT, a rolling streaming diffusion framework that preserves bounded bidirectional modeling within causal recurrent generation. Within a fixed-size window, LiveVVT jointly denoises multiple video chunks under bounded look-ahead, preserving local bidirectional interactions while emitting one clean chunk per iteration. Beyond the window, two complementary memories sustain long-term consistency: a bounded temporal memory propagates recent dynamics and occlusion context, whereas a persistent global appearance memory, constructed once from the target garment and a frontal try-on keyframe, anchors garment details and dressed appearance throughout the stream. We further introduce a progressive distillation framework integrating bidirectional VVT learning, teacher-trajectory regression for causal few-step adaptation, and Collaborative Matching Distillation, which couples teacher-distribution matching with rolling flow matching on real videos to align optimization with recurrent inference. Experiments on paired and unpaired long-sequence benchmarks demonstrate superior generation quality over similarly sized models, with $26\times$ lower latency and $11\times$ higher throughput, enabling high-fidelity real-time streaming VVT.
\end{abstract}

% !TeX root = ../main.tex

\section{Introduction}
\label{sec:intro}
Video Virtual Try-On (VVT)~\cite{li2025magictryon,zeng2025eevee,chong2025catv2ton} transfers a target garment onto a person throughout a video while preserving garment appearance, human identity, and temporal coherence. Unlike image-based try-on~\cite{chong2025catvton,zhang2026unifit,li2026dit}, VVT must resolve pose variation, occlusion, and nonrigid garment deformation consistently across frames \cite{zhong2021mv,jiang2022clothformer,zou2025video}. Its applications in online retail, digital humans, live-stream commerce, and augmented reality make high-fidelity real-time streaming a critical capability.

\begin{figure}[t]
\centering
\includegraphics[width=\linewidth]{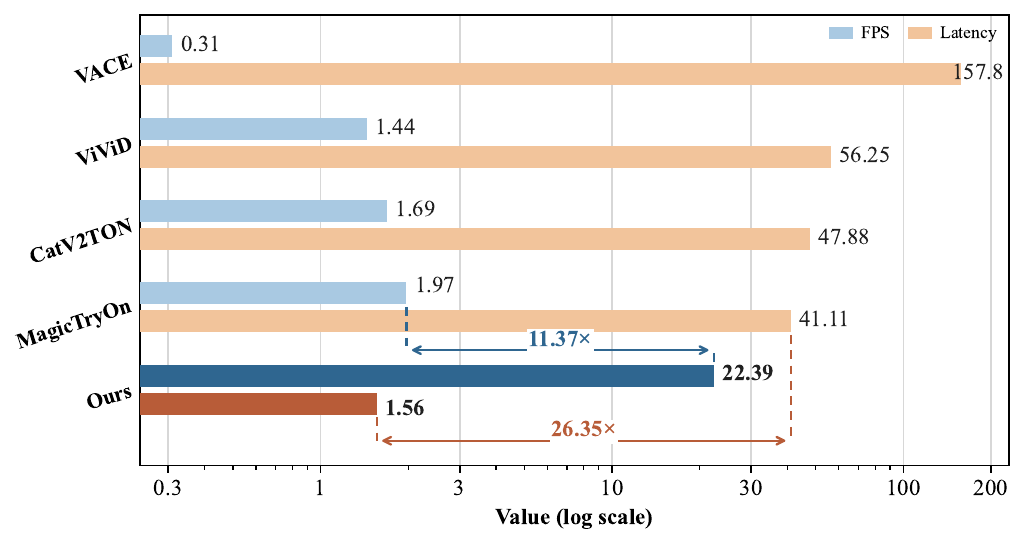}
\caption{Comparison of latency and throughput between LiveVVT and state-of-the-art methods.}
\label{fig:efficiency}
\end{figure}

Recent diffusion-based methods have substantially advanced VVT fidelity by integrating multimodal conditioning with spatio-temporal attention \cite{zou2025video,chong2025catv2ton,zuo2025dreamvvt}. However, the dominant paradigm remains clip-based and bidirectional: it depends on future frames and partitions long videos into independently processed windows, resulting in high latency and redundant computation. Although recent work explores interactive customization with arbitrary garments \cite{song2026fashionchameleon,zheng2026itryon}, continuous, source-conditioned streaming VVT that jointly delivers high fidelity and low latency remains largely unexplored.

High-fidelity streaming VVT cannot be achieved by simply imposing causal attention on an offline model. First, this conversion disrupts pretrained bidirectional spatio-temporal priors, leading to severe degradation in synthesis quality. Second, long-term consistency requires sufficient historical context; however, unbounded history is computationally impractical, whereas aggressive truncation induces appearance drift. Third, the transition from offline denoising to recurrent generation changes both attention patterns and sampling trajectories, while conditioning on self-generated history further introduces exposure bias and error accumulation \cite{causvid,selfforcing,cui2026selfforcingpp,rollingforcing}. Practical streaming VVT therefore requires the generation mechanism, memory architecture, and capability-transfer strategy to be redesigned jointly.

To address these challenges, we introduce LiveVVT, a rolling streaming diffusion framework that combines local bidirectional denoising with causal recurrence across windows. Within a fixed-size window, LiveVVT jointly denoises chunks at staggered noise levels under bounded look-ahead and emits one clean chunk per iteration. Beyond the window, two complementary memories sustain long-term consistency: a bounded temporal memory propagates recent dynamics and occlusion context, while a persistent global appearance memory anchors garment details and overall dressed appearance throughout the stream. To transfer high-fidelity offline priors to recurrent few-step generation, progressive distillation integrates bidirectional VVT learning, teacher-trajectory regression, and Collaborative Matching Distillation (CoMD). CoMD couples teacher-distribution matching with rolling flow matching on real videos under recurrent inference, narrowing the gap between offline supervision and streaming generation. Experiments on paired and unpaired long-sequence benchmarks show that LiveVVT outperforms similarly sized models with $26\times$ lower latency and $11\times$ higher throughput.

Our main contributions are summarized as follows:
\begin{itemize}
\item We introduce LiveVVT, a rolling generation paradigm that combines bidirectional joint denoising within a staggered-noise window with causal recurrence across windows, enabling high-fidelity real-time VVT under bounded look-ahead.

\item We design two complementary memories: a bounded temporal memory propagates dynamics and occlusion context, while a persistent global appearance memory preserves garment texture and dressed appearance over long streams without repeated reference encoding.

\item We develop a progressive distillation framework that transfers high-fidelity offline VVT priors to a few-step streaming generator; its CoMD stage couples teacher distribution matching with rolling flow matching on real videos to align training with recurrent inference.

\end{itemize}

\section{Related Work}
\subsection{Image and Video Virtual Try-On}
Image virtual try-on has evolved from explicit garment warping and composition \cite{han2018viton,han2019clothflow,yang2020towards,ge2021parser} toward a conditional diffusion paradigm that improves realism and fine-grained detail preservation \cite{zhu2023tryondiffusion,choi2024improving,xu2025ootdiffusion,cao2026multivariate}. Extending this task to video, early methods maintain temporal coherence through optical-flow-based propagation and explicit inter-frame correspondence \cite{dong2019fw,zhong2021mv,jiang2022clothformer}, while modern generative architectures jointly denoise clips using temporal attention and multimodal conditioning \cite{zou2025video,chong2025catv2ton,zuo2025dreamvvt,li2025magictryon,cao2026univvt}. More recent efforts explicitly model human--garment interactions \cite{zheng2026itryon} or support customization with arbitrary garments \cite{song2026fashionchameleon} under complex motion and occlusion. Nevertheless, most high-fidelity VVT models remain clip-based and bidirectional, requiring complete input clips and redundant windowed inference for long videos. In contrast, LiveVVT enables source-conditioned streaming try-on through bounded rolling denoising and two complementary memory mechanisms: a recurrent temporal memory and a persistent global appearance memory, thereby achieving low-latency, high-throughput real-time streaming inference.

\subsection{Autoregressive Video Generation}
Autoregressive video generation has evolved from discrete visual-token prediction \cite{videogpt,phenaki} to continuous latent-space modeling \cite{nova}. Recent studies transform bidirectional diffusion models into few-step causal generators through trajectory- or distribution-level distillation, self-generated rollouts, and adversarial post-training \cite{causvid,selfforcing,zhu2026causal,zhao2026causal,aapt}, thereby mitigating causal adaptation, exposure bias, and accumulated generation errors. Recurrent and rolling architectures further extend this paradigm to real-time, long-form, and interactively controlled synthesis \cite{streamdit,cui2026selfforcingpp,longlive,rollingforcing,shin2026motionstream}. These approaches, however, primarily synthesize unconstrained content from text prompts or sparse controls. Streaming VVT imposes substantially stricter conditioning requirements: every output chunk must remain precisely aligned with the incoming person stream while consistently preserving identity, garment fidelity, and long-term temporal coherence, even under challenging motion and occlusion. LiveVVT addresses this densely conditioned setting by coupling bounded-look-ahead generation with persistent memory mechanisms and explicitly aligning few-step distillation with rolling inference on real video sequences at deployment scale.

% !TEX root = ../main.tex
\section{Method}
\label{sec:methods}

\subsection{Problem Formulation and Overview}
\label{sec:overview}

\begin{figure*}[t]
\centering
\includegraphics[width=\textwidth]{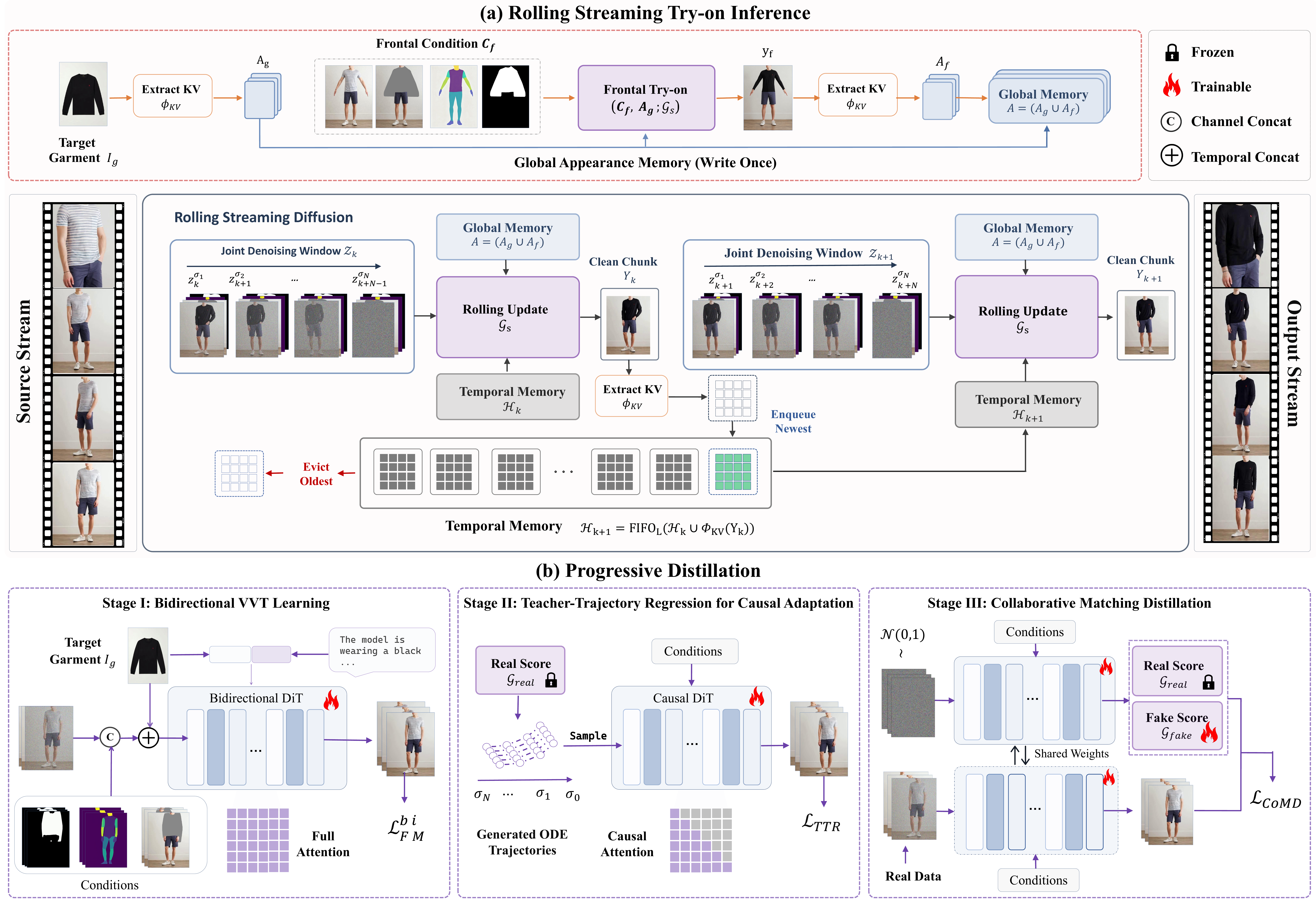}
\caption{Overview of LiveVVT. (a) Rolling streaming try-on emits one clean chunk per update from a staggered-noise window with persistent global appearance memory $\mathcal{A}=(\mathcal{A}_g,\mathcal{A}_f)$ and temporal memory $\mathcal{H}_k$. (b) Bidirectional VVT learning, teacher-trajectory regression, and CoMD progressively transfer offline bidirectional priors to few-step streaming generation.}
\label{fig:framework}
\end{figure*}

Let $\mathcal{X}=\{x_t\}_{t=1}^{T}$ denote the source-person stream and $I_g$ the target garment image. Each frame yields the try-on condition
\begin{equation}
c_t=(m_t,p_t,\bar{x}_t),
\end{equation}
where $m_t$, $p_t$, and $\bar{x}_t$ are the inpainting mask, DensePose map, and garment-agnostic person image. We partition the video and conditions into $K$ non-overlapping chunks $\{X_k\}_{k=1}^{K}$ and $\{C_k\}_{k=1}^{K}$. The garment condition is denoted by $\mathbf{f}_g$; $t$ indexes frames, $k$ indexes chunks, and $\tau\in[0,1]$ is reserved for normalized diffusion time.

LiveVVT introduces a frontal source keyframe $x_f$ and its condition $C_f$ to construct a persistent appearance anchor; $x_f$ is captured in a standard A-Pose at deployment. For training videos without such a reference, we select the frame whose pose is closest to the A-Pose template $p_{\mathrm a}$:
\begin{equation}
x_f=x_{\arg\min_{t\in\{1,\ldots,T\}}d(p_t,p_{\mathrm a})},
\label{eq:apose}
\end{equation}
where $d(\cdot,\cdot)$ is the pose-alignment distance.

At recurrent update $k$, the streaming generator produces one completed try-on chunk:
\begin{equation}
Y_k=\mathcal{G}_s(C_{k:k+N-1},\mathcal{H}_k,\mathcal{A}),
\label{eq:streaming_objective}
\end{equation}
where $N$ is the window size, $\mathcal{H}_k$ a bounded temporal memory, and $\mathcal{A}$ a persistent global appearance memory constructed from the garment and frontal reference. Figure~\ref{fig:framework} illustrates the LiveVVT framework. Under bounded look-ahead, a fixed-size rolling window preserves bidirectional interactions and emits one fully denoised chunk per update. Across windows, temporal memory recurrently propagates recent motion dynamics and occlusion context, while global appearance memory persistently anchors garment details and overall dressed appearance throughout the stream. To bridge the mismatch in attention patterns and sampling trajectories between bidirectional generation and causal recurrence, we further introduce a progressive distillation framework that progressively transfers the high-fidelity generative prior of an offline bidirectional model to a causal few-step generator.

\subsection{Rolling Streaming Try-On}
\label{sec:rolling}

Offline bidirectional diffusion jointly denoises a clip at a shared noise level, requiring both the full input and the complete sampling trajectory before emission. LiveVVT instead organizes denoising as a rolling pipeline. At each update, resident chunks advance by one stage and a fresh Gaussian-noise chunk enters the window; chunks of different ages therefore occupy staggered noise levels. Let $0=\sigma_0<\sigma_1<\cdots<\sigma_N=1$ denote an $N$-step schedule from clean data to Gaussian noise. Before update $k$, the active window is
\begin{equation}
\mathcal{Z}_k=(z_k^{\sigma_1},z_{k+1}^{\sigma_2},\ldots,z_{k+N-1}^{\sigma_N}).
\label{eq:mixed_noise_window}
\end{equation}
The leading chunk is one step from clean, whereas the newest remains pure noise. Given $\boldsymbol{\sigma}=(\sigma_1,\ldots,\sigma_N)$, each streaming update jointly advances all chunks from $\sigma_i$ to $\sigma_{i-1}$:
\begin{equation}
\bar{\mathcal{Z}}_k=
\operatorname{Step}\!\left(
\mathcal{Z}_k,
\mathcal{G}_s(\mathcal{Z}_k,\boldsymbol{\sigma},C_{k:k+N-1},\mathcal{H}_k,\mathcal{A},\mathbf{f}_g)
\right),
\label{eq:joint_step}
\end{equation}
producing
\begin{equation}
\bar{\mathcal{Z}}_k=(z_k^{\sigma_0},z_{k+1}^{\sigma_1},\ldots,z_{k+N-1}^{\sigma_{N-1}}).
\label{eq:window_denoised}
\end{equation}
Only the leading chunk reaches $\sigma_0$, so each update completes one chunk while refining the rest. Despite asynchronous noise levels, active chunks interact through bidirectional attention, allowing the leading chunk to exploit partially denoised future context for deformation and occlusion reasoning. Beyond the active window, $\mathcal{H}_k$ propagates generated context, while $\mathcal{A}$ provides persistent appearance cues; no future frames are accessed across updates. LiveVVT combines bounded-look-ahead bidirectional denoising within each window with causal recurrence across windows. The completed latent is decoded by the frozen VAE decoder $\mathcal{D}$:
\begin{equation}
Y_k=\mathcal{D}(z_k^{\sigma_0}).
\label{eq:decode_chunk}
\end{equation}

LiveVVT then removes the emitted latent, shifts the remaining chunks, and appends fresh noise:
\begin{equation}
\mathcal{Z}_{k+1}=(z_{k+1}^{\sigma_1},\ldots,z_{k+N-1}^{\sigma_{N-1}},z_{k+N}^{\sigma_N}),
\label{eq:rolling_update}
\end{equation}
where $z_{k+N}^{\sigma_N}\sim\mathcal{N}(0,I)$. This roll--append operation restores Eq.~\ref{eq:mixed_noise_window} for the next update: every chunk enters at $\sigma_N$, undergoes exactly $N$ joint updates, and exits at $\sigma_0$. Confining joint denoising to a fixed window makes per-update computation and activation memory independent of stream length.

\subsection{Dual-Memory Mechanism}
\label{sec:global_memory}

The bounded window cannot retain latent context after a chunk rolls out. This truncation induces two inconsistencies: motion and occlusion states may become discontinuous across adjacent windows, while garment details and dressed appearance may drift over longer horizons. LiveVVT addresses these failure modes with memories at two temporal scales: an evolving temporal memory for recent cross-window dynamics and a persistent global appearance memory for stable sequence-level anchoring.

\paragraph{Temporal memory.}
To maintain cross-window continuity, LiveVVT caches attention key--value (KV) features from each completed clean latent:
\begin{equation}
\mathcal{H}_{k+1}=\operatorname{FIFO}_{L}\!\left(
\mathcal{H}_{k}\cup\Phi_{\mathrm{KV}}(z_k^{\sigma_0})
\right),
\label{eq:temporal_cache}
\end{equation}
where $\operatorname{FIFO}_{L}$ retains the latest $L$ entries. Subsequent windows attend to the cached features, propagating recent motion states and occlusion relations without reprocessing emitted chunks. Because the cache is derived from generated clean latents, it matches the self-conditioned history encountered during recurrent inference. Its bounded capacity limits online cost but evicts older states; $\mathcal{H}_k$ is therefore suited to local dynamics rather than persistent appearance anchoring.

\paragraph{Global appearance memory.}
This limitation motivates a complementary reference for stable, person-specific dressed appearance over extended horizons. LiveVVT therefore constructs a persistent global appearance memory that anchors garment details and overall dressed appearance throughout the stream. We first extract garment KV features
\begin{equation}
\mathcal{A}_g=\Phi_{\mathrm{KV}}(I_g),
\label{eq:garment_anchor}
\end{equation}
which preserve source texture and pattern but do not specify how the garment should appear on the target body. The frontal keyframe provides a canonical person-specific configuration for establishing dressed appearance before streaming. We combine its condition $C_f$ with $\mathcal{A}_g$ to generate a frontal try-on reference
\begin{equation}
y_f=\operatorname{TryOn}(C_f,\mathcal{A}_g,\mathbf{f}_g;\mathcal{G}_s),
\label{eq:frontal_tryon}
\end{equation}
and extract the corresponding KV features:
\begin{equation}
\mathcal{A}_f=\Phi_{\mathrm{KV}}(y_f).
\label{eq:frontal_kv}
\end{equation}
$\mathcal{A}_g$ and $\mathcal{A}_f$ form the persistent global appearance memory
\begin{equation}
\mathcal{A}=\mathcal{A}_g\cup\mathcal{A}_f.
\label{eq:global_anchor}
\end{equation}
Computed once before streaming, $\mathcal{A}$ is reused by every rolling window as a time-invariant appearance anchor. It complements the evolving temporal memory $\mathcal{H}_k$: $\mathcal{H}_k$ propagates recent motion dynamics and occlusion states, whereas $\mathcal{A}$ anchors garment details and overall dressed appearance, suppressing long-term drift. This dual-memory design extends context beyond the active window without retaining unbounded history or repeatedly encoding reference images.

\subsection{Progressive Distillation Framework}
\label{sec:progressive_distillation}

Rolling inference alters the attention pattern and denoising trajectory of offline bidirectional generation. Joint adaptation lacks stable causal initialization and effective supervision for history-conditioned student states. We therefore transfer capabilities progressively: bidirectional VVT learning establishes high-fidelity try-on; teacher-trajectory regression adapts the student to causal few-step denoising; and Collaborative Matching Distillation aligns recurrent outputs with teacher and real-video distributions. Each stage resolves a distinct mismatch while retaining acquired capabilities.

\paragraph{Stage I: Bidirectional VVT learning.}
We initialize $\mathcal{G}_b$ from Wan2.1-Fun-V1.1-1.3B-Control~\cite{wan2025wan}. VAE-encoded try-on conditions are concatenated channel-wise with noisy video latents, while the garment latent is prepended temporally. CLIP visual~\cite{radford2021learning} and UMT5-XXL~\cite{chung2023unimax} text embeddings
\[
\mathbf{f}_g=\{\mathbf{f}_g^{\mathrm{vis}},\mathbf{f}_g^{\mathrm{txt}}\}
\]
provide garment semantics through cross-attention. We extend the DiT input projection for these conditions while retaining its bidirectional spatio-temporal Transformer.

For a clean latent $z_0$, Gaussian noise $\epsilon\sim\mathcal{N}(0,I)$, and $\tau\sim\mathcal{U}(0,1)$, Flow Matching~\cite{lipman2022flow} defines the linear path and target velocity as
\begin{equation}
z_\tau=(1-\tau)z_0+\tau\epsilon,
\label{eq:flow_path}
\end{equation}
\begin{equation}
u_\tau=\epsilon-z_0,
\label{eq:target_velocity}
\end{equation}
We optimize $\mathcal{G}_b$ with
\begin{equation}
\mathcal{L}_{\mathrm{FM}}^{\mathrm{bi}}=
\mathbb{E}\!\left[
\left\|\mathcal{G}_b(z_\tau,\tau,C_{1:K},\mathbf{f}_g)-u_\tau\right\|_2^2
\right].
\label{eq:flow_matching}
\end{equation}
This stage preserves the bidirectional video prior and provides a high-fidelity initialization for $\mathcal{G}_s$. Applying the objective to Wan2.1-Fun-V1.1-14B-Control~\cite{wan2025wan} yields the teacher $\mathcal{G}_{\mathrm{real}}$ as the real-score model for Collaborative Matching Distillation.

\paragraph{Stage II: Causal adaptation.}
We initialize $\mathcal{G}_s$ from $\mathcal{G}_b$ to retain high-fidelity VVT while adapting it to causal few-step sampling. Following \cite{yin2025slow}, we use $\mathcal{G}_{\mathrm{real}}$ and an ODE solver to precompute $6{,}000$ teacher trajectories, each subsampled at the student's $N$-step schedule. Given an intermediate state $z_{\tau_j}^{i}$ and endpoint $z_0^i$ from trajectory $i$, teacher-trajectory regression predicts the endpoint:
\begin{equation}
\mathcal{L}_{\mathrm{TTR}}=
\mathbb{E}_{i,j}\!\left[
\left\|\mathcal{G}_s(z_{\tau_j}^{i},\tau_j,C^i)-z_0^i\right\|_2^2
\right].
\label{eq:trajectory_regression}
\end{equation}
This objective jointly adapts the student's attention pattern and denoising trajectory to causal few-step inference, providing a stable initialization for distribution-level distillation.

\paragraph{Stage III: Collaborative matching distillation.}
Teacher-trajectory regression adapts isolated sampling states but leaves the recurrent rollout distribution unconstrained. DMD~\cite{yin2024improved} narrows this gap using the real-score model $\mathcal{G}_{\mathrm{real}}$ and auxiliary fake-score model $\mathcal{G}_{\mathrm{fake}}$, but remains teacher-driven and lacks direct supervision of real-video states under rolling deployment. We therefore propose CoMD, coupling teacher-distribution matching with Rolling Flow Matching (RFM) on real videos:
\begin{equation}
\mathcal{L}_{\mathrm{CoMD}}=
\mathcal{L}_{\mathrm{DMD}}+\lambda\cdot\mathcal{L}_{\mathrm{RFM}}.
\label{eq:comd}
\end{equation}
RFM samples a ground-truth window and builds its prefix cache $\mathcal{H}_k$ and global appearance memory $\mathcal{A}$ in the rolling configuration. Instead of perturbing the window at a shared continuous time $\tau$, it assigns successive chunks the ordered vector $\boldsymbol{\sigma}=(\sigma_1,\ldots,\sigma_N)$ from the student's discrete schedule, reproducing inference-time staggered noise. Applying Eq.~\ref{eq:flow_matching} then supervises the real-data velocity field under deployment-matched memory and noise states. During alternating optimization, $\mathcal{G}_{\mathrm{fake}}$ learns the score of detached student rollouts, while $\mathcal{G}_s$ combines the DMD distribution gradient with RFM supervision on real windows. DMD preserves the teacher's generative prior, whereas RFM anchors recurrent outputs to the real-video distribution. Together, they close the teacher--student distribution gap and training--inference gap from self-generated history, mitigating recurrent exposure bias. Algorithm~\ref{alg:comd} summarizes the procedure.

\begin{algorithm}[t]
\caption{Collaborative Matching Distillation}
\label{alg:comd}
\textbf{Input}: dataset $\mathcal{D}$; student $\mathcal{G}_s$; real score $\mathcal{G}_{\mathrm{real}}$; schedule $\mathcal{T}_N=\{\sigma_i\}_{i=0}^{N}$; weight $\lambda$; update interval $r>1$ \newline
\textbf{Output}: streaming try-on model $\mathcal{G}_s$
\begin{algorithmic}[1]
\STATE $\mathcal{G}_{\mathrm{fake}}\leftarrow\operatorname{Copy}(\mathcal{G}_s)$; $s\leftarrow1$
\WHILE{\emph{not converged}}
    \STATE Sample $(X_{1:K},C_{1:K},I_g,C_f,\mathbf{f}_g)\sim\mathcal{D}$
    \STATE $\mathcal{A}\leftarrow\operatorname{BuildGlobalMemory}(I_g,C_f,\mathbf{f}_g)$
    \IF{$s\bmod r=0$}
        \STATE ${\mathcal{Z}}^{\sigma_0}_{1:K}\leftarrow\operatorname{RollingSample}(\mathcal{G}_s,C_{1:K},\mathcal{A},\mathbf{f}_g,\mathcal{T}_N)$
        \STATE $\mathcal{L}_{\mathrm{DMD}}\leftarrow\operatorname{DMD}({\mathcal{Z}}^{\sigma_0}_{1:K},\mathcal{G}_{\mathrm{real}},\mathcal{G}_{\mathrm{fake}},C_{1:K},I_g,\mathbf{f}_g)$
        \STATE Sample a window $X_{k:k+N-1}$; encoded as $\mathcal{Z}_{k}=(z_k,\ldots,z_{k+N-1})$ and build $\mathcal{H}_k$
        \STATE Set $\boldsymbol{\sigma}=(\sigma_1,\ldots,\sigma_N)$ and sample $\epsilon\sim\mathcal{N}(0,I)$
        \STATE $\mathcal{Z}_{\boldsymbol{\sigma},k}\leftarrow\{(1-\sigma_i)z_{k+i-1}+\sigma_i\epsilon_i\}_{i=1}^{N}$
        \STATE $\mathcal{L}_{\mathrm{RFM}}\leftarrow\|\mathcal{G}_s(\mathcal{Z}_{\boldsymbol{\sigma},k},\boldsymbol{\sigma},C_{k:k+N-1},\mathcal{H}_k,\mathcal{A},\mathbf{f}_g)-(\epsilon-\mathcal{Z}_{k})\|_2^2$
        \STATE $\mathcal{L}_{\mathrm{CoMD}}\leftarrow\mathcal{L}_{\mathrm{DMD}}+\lambda\cdot\mathcal{L}_{\mathrm{RFM}}$
        \STATE $\mathcal{G}_s\leftarrow\operatorname{Update}(\mathcal{G}_s,\mathcal{L}_{\mathrm{CoMD}})$
    \ELSE
        \STATE ${\mathcal{Z}}^{\sigma_0}_{1:K}\leftarrow\operatorname{RollingSample}(\mathcal{G}_s,C_{1:K},\mathcal{A},\mathbf{f}_g,\mathcal{T}_N)$
        \STATE ${\mathcal{Z}}^{\sigma_0}_{1:K}\leftarrow\operatorname{StopGrad}({\mathcal{Z}}^{\sigma_0}_{1:K})$
        \STATE Sample $\tau\sim\mathcal{U}(0,1)$ and $\epsilon\sim\mathcal{N}(0,I)$
        \STATE $\mathcal{Z}^\tau_{1:K}\leftarrow(1-\tau)\mathcal{Z}^{\sigma_0}_{1:K}+\tau\epsilon$
        \STATE $\mathcal{L}_{\mathrm{fake}}\leftarrow\|\mathcal{G}_{\mathrm{fake}}(\mathcal{Z}^\tau_{1:K},\tau,C_{1:K},I_g,\mathbf{f}_g)-(\epsilon-\mathcal{Z}^{\sigma_0}_{1:K})\|_2^2$
        \STATE $\mathcal{G}_{\mathrm{fake}}\leftarrow\operatorname{Update}(\mathcal{G}_{\mathrm{fake}},\mathcal{L}_{\mathrm{fake}})$
    \ENDIF
    \STATE $s\leftarrow s+1$
\ENDWHILE
\end{algorithmic}
\end{algorithm}

% !TeX root = ../main.tex

\section{Experiments}
\label{sec:expriments}
\begin{table}[t]
    \centering
    \footnotesize
    \renewcommand{\arraystretch}{0.8}
    \setlength{\tabcolsep}{0.6pt}
    \begin{tabular}{@{}lrrrrrr@{}}
        \toprule
        & \multicolumn{4}{c}{Paired} & \multicolumn{2}{c}{Unpaired} \\
        \cmidrule(lr){2-5}\cmidrule(l){6-7}
        Method
        & VFID$_I$$\downarrow$
        & VFID$_R$$\downarrow$
        & SSIM$\uparrow$
        & LPIPS$\downarrow$
        & VFID$_I$$\downarrow$
        & VFID$_R$$\downarrow$ \\
        \midrule
        OOTDiff+AM       & 46.666 & 24.450 & 0.720 & 0.238  & 47.802 & 25.857 \\
        CatVTON+AM       & 45.754 & 20.707 & 0.739 & 0.231  & 46.125 & 20.728 \\
        VACE  & 18.086 &  0.418 & 0.762 & 0.149  & 27.519 &  0.871 \\
        ViViD            & 20.887 &  0.337 & 0.792 & 0.130  & 27.963 &  0.462 \\
        CatV2TON         & 20.694 &  0.232 & 0.815 & 0.132  & 28.343 &  0.356 \\
        MagicTryOn  & \underline{16.148} &  \underline{0.228} & \underline{0.835} & \textbf{0.091}  & \underline{24.137} &  \underline{0.287} \\
        \addlinespace[1pt]
        \rowcolor{blue!5}
        \textbf{Ours}   & \textbf{13.194} & \textbf{0.090} & \textbf{0.838} & \underline{0.099} & \textbf{23.608} &  \textbf{0.213} \\
        \bottomrule
    \end{tabular}
    \caption{Quantitative comparison of LiveVVT with competing methods on the long-sequence ViViD-SL benchmark.}
    \label{tab:vivid-sl}
    \label{tab}
\end{table}

\begin{table}[t]
    \centering
    \footnotesize
    \renewcommand{\arraystretch}{0.8}
    \setlength{\tabcolsep}{0.6pt}
    \begin{tabular}{@{}lrrrrrr@{}}
        \toprule
        & \multicolumn{4}{c}{Paired} & \multicolumn{2}{c}{Unpaired} \\
        \cmidrule(lr){2-5}\cmidrule(l){6-7}
        Method
        & VFID$_I$$\downarrow$
        & VFID$_R$$\downarrow$
        & SSIM$\uparrow$
        & LPIPS$\downarrow$
        & VFID$_I$$\downarrow$
        & VFID$_R$$\downarrow$ \\
        \midrule
        OOTDiff+AM       & 33.225 & 2.311 & 0.756 & 0.246 & 39.139 & 2.172 \\
        CatVTON+AM       & 33.334 & 2.555 & 0.769 & 0.246 & 36.863 & 1.547 \\
        VACE  & 20.373 & 0.405 & 0.787 & 0.173 & 31.152 & 1.648 \\
        ViViD            & 24.244 & 0.611 & 0.785 & 0.145 & 28.356 & 1.037 \\
        CatV2TON         & 22.018 & 0.376 & \underline{0.811} & 0.156  & 30.653 & 1.133   \\
        MagicTryOn  & \underline{17.988} & \underline{0.370} & 0.808 & \textbf{0.112} & \underline{25.950} & \textbf{0.832} \\
        \addlinespace[1pt]
        \rowcolor{blue!5}
        \textbf{Ours}   & \textbf{15.020} & \textbf{0.290} & \textbf{0.814} & \underline{0.115} & \textbf{24.297}  & \underline{1.024}    \\
        \bottomrule
    \end{tabular}
    \caption{Quantitative comparison of LiveVVT with competing methods on the long-sequence ViT-HDL benchmark.}
    \label{tab:vithd-l}
\end{table}

\subsection{Experimental Setup}
\paragraph{Datasets.}
We train on two image datasets (VITON-HD~\cite{choi2021viton} and DressCode\cite{morelli2022dress}) and three video datasets (ViViD~\cite{fang2024vivid}, ViT-HD~\cite{He_2026_CVPR}, and TikTokDress~\cite{nguyen2025swifttry}), yielding 60,039 paired image samples and 23,258 paired video samples after preprocessing and filtering. Existing VVT studies commonly evaluate performance on subsets of 64-frame frontal-view clips, which provide limited evidence of long-term consistency. We therefore construct ViViD-SL from the 60 longest videos in the independent ViViD-S test set and ViT-HDL from 60 longest videos in the ViT-HD test split. The resulting sequences span 194--420 frames; all frames are evaluated under both paired and unpaired settings.

\paragraph{Evaluation Metrics.}
We assess video-level quality using VFID with I3D~~\cite{carreira2017quo} and ResNeXt~\cite{xie2017aggregated} feature extractors, denoted as VFID$_I$ and VFID$_R$, respectively. SSIM and LPIPS measure frame-level structural similarity and perceptual distance to the ground truth. Lower VFID and LPIPS and higher SSIM indicate better performance. We report all four metrics for paired evaluation; for unpaired evaluation, where frame-wise ground truth is unavailable, we report VFID$_I$ and VFID$_R$.

\paragraph{Implementation Details.}
We train LiveVVT on eight NVIDIA A100 (80 GB) GPUs using AdamW with a batch size of 8 and weight decay of 0.01. Bidirectional VVT learning runs for 30k steps at a learning rate of $1\times10^{-5}$, followed by 20k steps of teacher-trajectory regression at $2\times10^{-6}$. CoMD is performed for 20k steps, using learning rates of $1.5\times10^{-6}$ and $4\times10^{-7}$ for the student generator and auxiliary fake-score model, respectively. We use a rolling window of $N=4$ chunks with three latent frames per chunk and set the RFM weight to $\lambda=0.2$. The temporal-memory cache size is set to L=21, following Self-Forcing \cite{selfforcing}. Training and inference use a resolution of $512\times384$, and evaluations use a fixed random seed of 42.
%需要交待窗口LiveVVT最终采用的窗口大小N和去噪步数，每个chunk包含的latent frames数

\subsection{Quantitative Experiments}
We compare LiveVVT against the two-stage image-to-video pipelines\footnote{These pipelines first synthesize a try-on result for the initial frame using OOTDiffusion or CatVTON, and then animate it with AnimateAnyone conditioned on the source pose sequence.} OOTDiff+AM~\cite{xu2025ootdiffusion,hu2024animate} and CatVTON+AM~\cite{chong2025catvton}; the offline bidirectional VVT models ViViD~\cite{fang2024vivid}, CatV2TON~\cite{chong2025catv2ton}, and MagicTryOn\footnote{We use MagicTryOn-1.3B throughout this work, as its parameter count is comparable to that of LiveVVT.}~\cite{li2025magictryon}; and VACE~\cite{jiang2025vace}, a unified video generation and editing model. To respect the clip-based training configurations of the competing methods and ensure fair comparison under GPU memory constraints, all baselines process long videos using 81-frame windows, whereas LiveVVT performs native rolling inference over the stream.

\paragraph{Long-sequence try-on quality.}
Tables~\ref{tab:vivid-sl} and~\ref{tab:vithd-l} evaluate long-sequence try-on fidelity. The two-stage pipelines consistently underperform video-native methods, indicating that animating a single try-on frame cannot adequately capture evolving garment deformation, occlusion, and person--garment interactions. Despite the constraints of causal recurrence, bounded look-ahead, and few-step sampling, LiveVVT delivers the strongest video quality. On ViViD-SL, it ranks first in paired VFID$_I$, VFID$_R$, and SSIM as well as both unpaired VFID metrics, while retaining competitive LPIPS. On ViT-HDL, LiveVVT again achieves the best paired VFID and SSIM scores and the best unpaired VFID$_I$. These results show that confining bidirectional denoising to a fixed-size rolling window does not compromise long-video fidelity. Local bidirectional interactions capture short-term garment deformation, while the temporal and global appearance memories propagate dynamic context and appearance constraints beyond the active window. Together, these components preserve garment identity, structural integrity, and temporal stability throughout long video streams.
\begin{figure}[t]
\centering
\includegraphics[width=\linewidth]{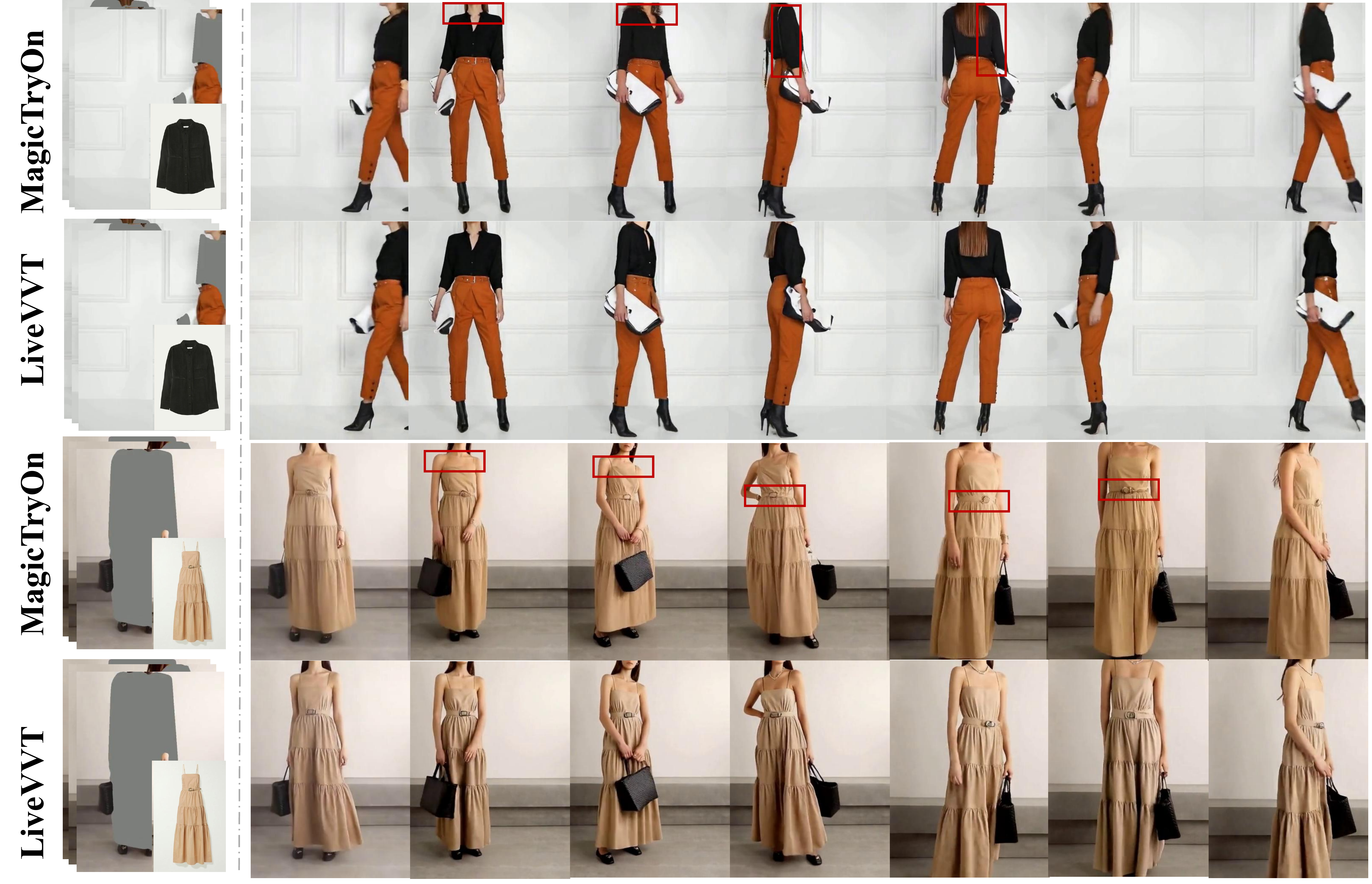}
\caption{Long-Term and Multi-View Try-On Comparison.}
\label{fig:visual_comparison}
\end{figure}
\paragraph{Real-time efficiency.}
Figure~\ref{fig:efficiency} compares first-chunk latency and sustained throughput at $512\times384$ resolution. LiveVVT produces its first chunk in $1.56$ seconds and subsequently sustains $22.39$ FPS. First-chunk latency determines perceived responsiveness: a shorter delay allows users to view the try-on result without waiting for the complete source clip or lengthy offline processing, while sustained throughput governs the continuity of subsequent updates. Relative to the similarly sized MagicTryOn, LiveVVT reduces latency by $26.35\times$ and improves throughput by $11.37\times$; against the substantially larger VACE, these gains reach $100.64\times$ and $72.22\times$, respectively. Together with the preceding quality results, these efficiency gains establish LiveVVT not as a merely accelerated offline model, but as a genuinely high-fidelity streaming video virtual try-on system.

\begin{figure*}[t]
    \centering
    \begin{minipage}[t]{0.3\textwidth}
        \centering
        \includegraphics[width=\linewidth]{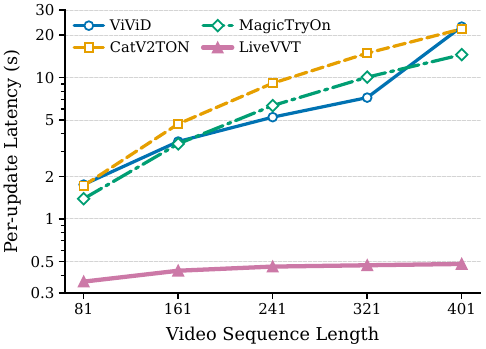}
        \small (a) Per-update latency
    \end{minipage}
    \hfill
    \begin{minipage}[t]{0.3\textwidth}
        \centering
        \includegraphics[width=\linewidth]{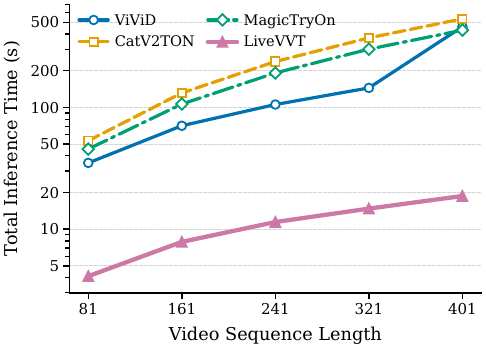}
        \small (b) Total inference time
    \end{minipage}
    \hfill
    \begin{minipage}[t]{0.3\textwidth}
        \centering
        \includegraphics[width=\linewidth]{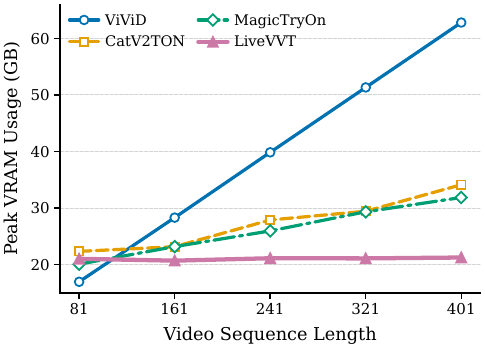}
        \small (c) Peak GPU memory
    \end{minipage}
    \caption{Long-stream scalability: (a) per-update latency, (b) total inference time, and (c) peak GPU memory usage.}
    \label{fig:per_update_time_vram}
\end{figure*}

\paragraph{Long-stream scalability.}
Figure~\ref{fig:per_update_time_vram} profiles inference cost as sequence length increases. Offline bidirectional methods incur rapidly growing per-update and end-to-end latency because longer sequences require increasingly expensive joint spatio-temporal computation (Figures~\ref{fig:per_update_time_vram}(a,b)). In contrast, LiveVVT confines joint denoising to a fixed-size rolling window. Its per-update latency stabilizes at approximately $0.5$ seconds, while total inference time increases smoothly with stream duration. Its peak GPU memory is also nearly invariant to sequence length (Figure~\ref{fig:per_update_time_vram}(c)), since both the active denoising window and temporal KV cache are bounded and the global appearance memory is constructed only once. This length-decoupled update cost and memory footprint directly reflect the proposed rolling formulation, enabling continuous long-video try-on on resource-constrained hardware.

\subsection{Qualitative Experiments}
To examine long-horizon consistency beyond aggregate metrics, Figure~\ref{fig:visual_comparison} compares LiveVVT with MagicTryOn, the closest dedicated VVT competitor in quantitative performance. MagicTryOn preserves garment structure and texture within each window, but its clip-based formulation necessitates windowed inference on long sequences, causing appearance drift across window boundaries that becomes pronounced under large viewpoint changes. This drift is evident in the shoulder-bag strap in the first sequence and in the dress straps and waist ornament in the second. LiveVVT instead maintains garment geometry, fine-grained textures, and overall dressed appearance throughout both sequences. Its persistent global appearance memory provides a sequence-level appearance anchor, while recurrent rolling-window generation preserves continuity across updates; together, they suppress local prediction drift over extended horizons. Additional comparisons are provided in the appendix.

\subsection{Ablation Study}
\paragraph{Complementary memory mechanisms.}
Table~\ref{tab:dual-memory} isolates the contributions of the global memory's frontal try-on component $\mathcal{A}_f$ and temporal memory $\mathcal{H}$. All variants retain the garment component $\mathcal{A}_g$ because removing this essential target-garment condition would alter the task rather than test the memory design. Adding $\mathcal{A}_f$ reduces VFID$_I$ and VFID$_R$, demonstrating that a persistent dressed-person reference mitigates appearance drift over long sequences. The temporal memory $\mathcal{H}$ yields broader gains in video-level quality and frame-level consistency by propagating recent dynamics and occlusion states beyond the rolling window, improving cross-window continuity and structural coherence. Combining $\mathcal{H}$ and $\mathcal{A}_f$ achieves the best VFID$_I$, VFID$_R$, and SSIM, confirming their complementarity: $\mathcal{H}$ preserves local dynamics, whereas $\mathcal{A}_f$ anchors long-term dressed appearance.

\begin{figure}[t]
\centering
\includegraphics[width=\linewidth]{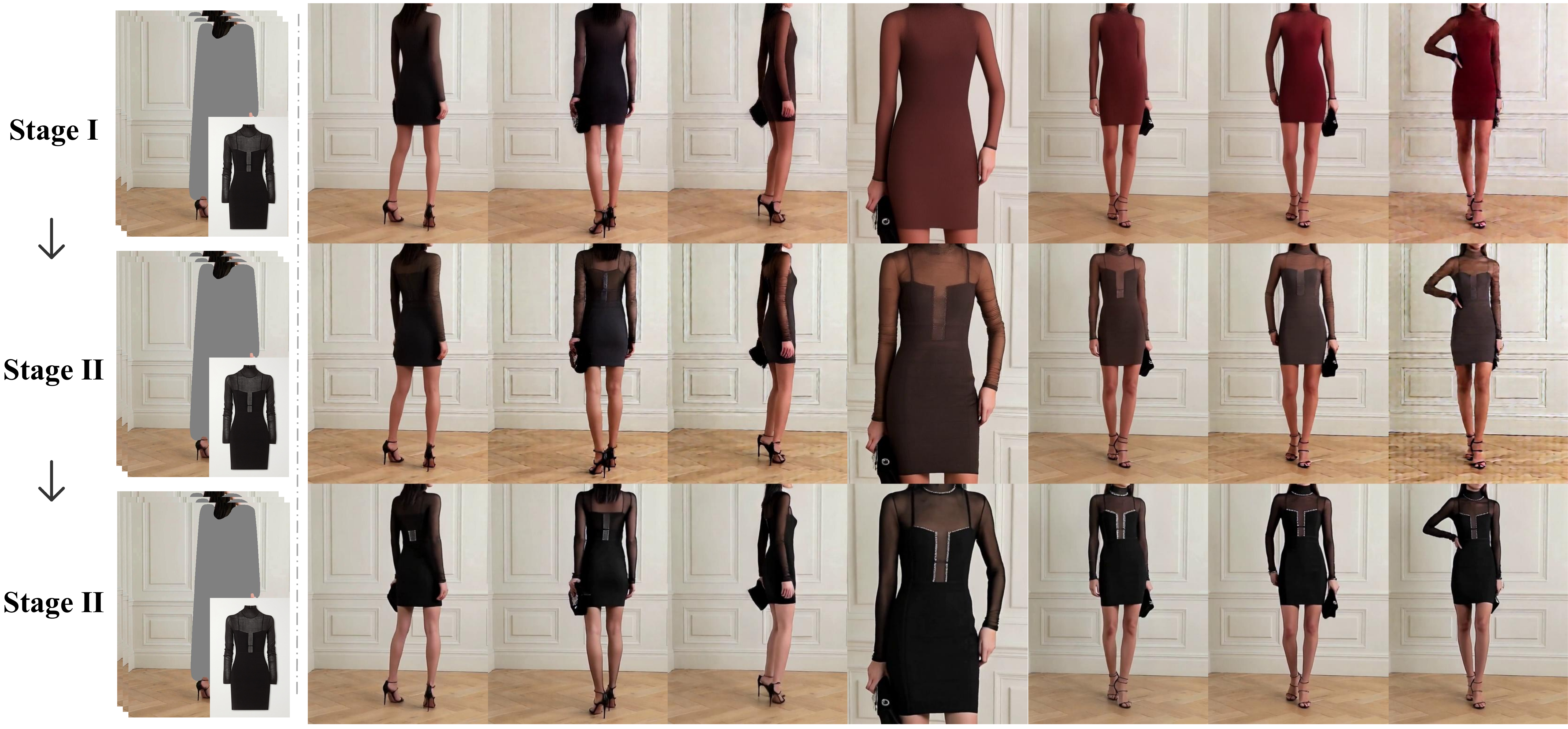}
\caption{Qualitative comparison across progressive distillation stages under rolling inference.}
\label{fig:progressive-training}
\end{figure}
\begin{table}[t]
    \centering
    \footnotesize
    \renewcommand{\arraystretch}{0.8}
    \setlength{\tabcolsep}{1.2pt}
    \begin{tabular}{@{}lrrrrr@{}}
        \toprule
        Configuration
        & VFID$_I$$\downarrow$
        & VFID$_R$$\downarrow$
        & SSIM$\uparrow$
        & LPIPS$\downarrow$
        & FPS$\uparrow$ \\
        \midrule
        $\mathcal{A}_g$ only                    & 14.415 & 0.131 & 0.824 & 0.094 & 26.74 \\
        $+\,\mathcal{A}_f$                        & 13.960 & 0.115 & 0.826 & 0.093 & 25.85 \\
        $+\,\mathcal{H}$                        & 13.871 & 0.112 & 0.832 & 0.088 & 22.60 \\
        $+\,\mathcal{H},\mathcal{A}_f$ (Ours) & 13.194 & 0.090 & 0.838 & 0.099 & 22.39 \\
        \bottomrule
    \end{tabular}
    \caption{Ablation of the frontal try-on and temporal memory components on ViViD-SL under paired setting.}
    \label{tab:dual-memory}
\end{table}

\begin{table}[t]
    \centering
    \footnotesize
    \renewcommand{\arraystretch}{0.8}
    \setlength{\tabcolsep}{4.0pt}
    \begin{tabular}{@{}lrrrr@{}}
        \toprule
        Training stage
        & VFID$_I$$\downarrow$
        & VFID$_R$$\downarrow$
        & SSIM$\uparrow$
        & LPIPS$\downarrow$ \\
        \midrule
        Stage I   & 17.202 & 0.281 & 0.786 & 0.144 \\
        Stage II  & 14.371 & 0.156 & 0.803 & 0.116 \\
        Stage III & 13.194 & 0.090 & 0.838 & 0.099  \\
        \bottomrule
    \end{tabular}
    \caption{Ablation of the three progressive distillation stages under rolling inference on ViViD-SL under paired setting.}
    \label{tab:progressive-training}
\end{table}

\paragraph{Progressive distillation.}
Table~\ref{tab:progressive-training} evaluates the three training stages under rolling denoising. Applying bounded-window recurrent inference directly to Stage I causes substantial degradation, as its altered attention and few-step trajectory disrupt the pretrained bidirectional prior. Stage II uses teacher-trajectory regression to adapt the student to causal attention and few-step sampling, improving metrics and providing a stable initialization for $\mathcal{G}_s$. Stage III applies CoMD to jointly match the teacher distribution and real-video rolling trajectories, yielding further gains across all metrics. Figure~\ref{fig:progressive-training} confirms this progression: Stage I shows structural distortion and texture drift; Stage II restores garment structure but retains color and texture discrepancies; Stage III accurately preserves garment geometry, appearance details, and long-term temporal consistency. These results validate progressive distillation for transferring offline bidirectional priors to high-fidelity streaming generation.

% \paragraph{Setting of Hyperparameter $\lambda$.}

% !TeX root = ../main.tex

\section{Conclusion}
\label{sec:conclusion}

We presented LiveVVT, a rolling diffusion framework for high-fidelity streaming VVT. A fixed staggered-noise window preserves local bidirectional spatio-temporal modeling under bounded look-ahead and emits one clean chunk per update. Complementary temporal and persistent global appearance memories propagate recent motion and occlusion context while anchoring garment details and dressed appearance across long streams. Progressive distillation integrates bidirectional VVT learning, teacher-trajectory regression, and CoMD to transfer offline bidirectional priors into causal few-step inference while aligning recurrent generation with teacher and real-video distributions. Experiments on paired and unpaired long-sequence benchmarks demonstrate strong fidelity, $26\times$ lower latency, and $11\times$ higher throughput than a similarly sized baseline, with per-update cost and peak memory nearly independent of stream length. Together, these results establish LiveVVT as a practical streaming VVT system and provide a path for adapting high-fidelity bidirectional video diffusion models to efficient online generation in densely conditioned video tasks.

% \newpage

\bibliography{aaai2027}
\newpage
\newpage
\appendix
\section{Additional Quantitative Comparisons}

To complement the long-sequence evaluations on ViViD-SL and ViT-HDL reported in the main paper, we introduce a longer benchmark, TikTokDress-L. We construct it by selecting the 60 longest videos from the TikTokDress test split~\cite{nguyen2025swifttry}. The resulting sequences, sourced from real-world TikTok videos, span 352--693 frames and contain diverse human motions, large viewpoint changes, and multiple garment categories. TikTokDress-L extends the evaluation to longer and more unconstrained sequences, where structural errors, garment-detail degradation, and cross-window appearance drift are more likely to accumulate during extended real-world streaming try-on.

Table~\ref{tab:tiktokdress-l} reports full-sequence results under paired and unpaired settings. LiveVVT achieves the best paired performance across all four metrics, with a VFID$_I$ of 19.668, a VFID$_R$ of 0.412, an SSIM of 0.842, and an LPIPS of 0.103. The strong VFID results demonstrate high video-level realism over hundreds of frames, while the simultaneous gains in SSIM and LPIPS show that this temporal stability is achieved without sacrificing frame-level garment structure or perceptual fidelity. In particular, LiveVVT remains robust under large pose and viewpoint changes, where independently processed clips are prone to inconsistent garment geometry, texture drift, and discontinuities at window boundaries.

LiveVVT also ranks first on both unpaired metrics, obtaining a VFID$_I$ of 30.447 and a VFID$_R$ of 0.762. The unpaired setting evaluates generalization to arbitrary target garments that are not matched to the source person's original clothing, more closely reflecting practical try-on scenarios. Since no frame-wise ground truth exists for these novel person--garment combinations, we assess their video-level realism and distributional consistency using VFID. The consistent performance across paired and unpaired settings provides strong evidence that LiveVVT preserves garment identity and dressed appearance throughout long, unconstrained sequences. These findings are consistent with the proposed rolling formulation: local bidirectional denoising models short-range deformation and motion within each active window, while temporal and persistent global appearance memories propagate dynamic context and appearance constraints beyond the window. Together, they enable stable long-horizon try-on generation without requiring full-sequence bidirectional inference.

\section{Fast Image Try-On}

Across all stages of progressive distillation, LiveVVT is trained on a multimodal dataset comprising dedicated video data together with image try-on data from VITON-HD~\cite{choi2021viton} and DressCode~\cite{morelli2022dress}. This unified training protocol enables the same model to perform both streaming video try-on and fast single-image try-on without a task-specific image branch. An image sample is represented as a one-frame video during training. Because no frontal keyframe is required in this setting, the global appearance memory contains only the garment attention key/value (KV) features, i.e., $\mathcal{A}=\mathcal{A}_g$. At inference, the image is processed with a single chunk ($N=1$) using the same few-step denoising interface as streaming generation.
\begin{table}[tb]
    \centering
    \footnotesize
    \renewcommand{\arraystretch}{0.8}
    \setlength{\tabcolsep}{0.6pt}
    \begin{tabular}{@{}lrrrrrr@{}}
        \toprule
         & \multicolumn{4}{c}{Paired} & \multicolumn{2}{c}{Unpaired} \\
        \cmidrule(lr){2-5}\cmidrule(lr){6-7}
        Method
        & VFID$_I$$\downarrow$
        & VFID$_R$$\downarrow$
        & SSIM$\uparrow$
        & LPIPS$\downarrow$
        & VFID$_I$$\downarrow$
        & VFID$_R$$\downarrow$ \\
        \midrule
        OOTDiff+AM       & 42.955 & 1.816 & 0.656 & 0.259 & 47.001 & 2.622 \\
        CatVTON+AM       & 33.043 & 1.422 & 0.714 & 0.213 & 39.726 & 1.192 \\
        VACE             & 23.589 & 0.875 & \underline{0.835} & 0.128 & \underline{32.292} & 2.378 \\
        ViViD           &  28.712 & \underline{0.812} & 0.791 & 0.149 & 35.954 & 1.316 \\
        CatV2TON         & 29.888 & 1.633 & 0.796 & 0.163 & 37.385 & 1.420 \\
        MagicTryOn  & \underline{23.007} & 0.927 & 0.818 & \underline{0.118} & 32.547 & \underline{1.043} \\
        \addlinespace[1pt]
        \rowcolor{blue!5}
        \textbf{Ours}   & \textbf{19.668} & \textbf{0.412} & \textbf{0.842} & \textbf{0.103} & \textbf{30.447} & \textbf{0.762} \\
        \bottomrule
    \end{tabular}
    \caption{Full-sequence quantitative comparison on the long-sequence TikTokDress-L benchmark under paired and unpaired settings.}
    \label{tab:tiktokdress-l}
\end{table}
Table~\ref{tab:zalando-test} evaluates this image capability on the 2,032-image VITON-HD test set at $512\times384$ resolution. FID and KID measure distributional fidelity, while SSIM and LPIPS quantify paired structural similarity and perceptual distance. LiveVVT obtains the best paired FID (5.910) and KID (0.419), together with the lowest paired LPIPS (0.054). Its paired SSIM (0.873) is close to the strongest result (0.883). Under the unpaired setting, which tests person--garment combinations without frame-wise correspondence, LiveVVT achieves the best KID (0.889) and the second-best FID (9.212). Taken together, these results show that the unified model preserves strong image try-on quality while sharing the same parameters and conditioning pathway with the streaming video model. The comparison does not by itself disentangle the individual effects of video and image supervision; rather, it verifies that multimodal training supports both capabilities within one model.

\begin{figure}[t]
\centering
\includegraphics[width=\linewidth]{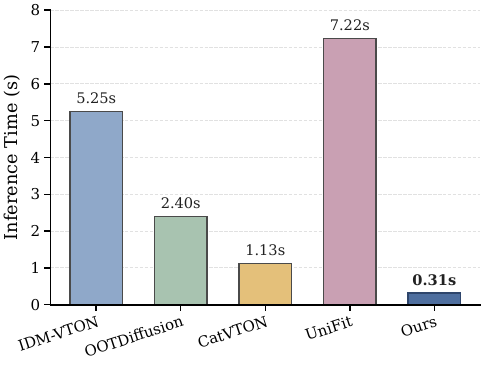}
\caption{Average single-image try-on inference time of LiveVVT and state-of-the-art image try-on methods, evaluated at a resolution of $512\times384$.}
\label{fig:vton_efficiency}
\end{figure}

Figure~\ref{fig:vton_efficiency} compares average single-image inference time at $512\times384$ resolution. LiveVVT requires 0.31 seconds per image, compared with 5.25, 2.40, 1.13, and 7.22 seconds for IDM-VTON~\cite{choi2024improving}, OOTDiffusion~\cite{xu2025ootdiffusion}, CatVTON~\cite{chong2025catvton}, and UniFit~\cite{zhang2026unifit}, respectively. These measurements correspond to speedups of $16.94\times$, $7.74\times$, $3.65\times$, and $23.29\times$. UniFit has the largest computational footprint because it combines a 12B-parameter FLUX.1-Fill-dev~\cite{labs2025flux1kontextflowmatching} backbone with a 2B-parameter auxiliary MLLM~\cite{Qwen2.5-VL} for semantic alignment. LiveVVT's efficiency comes from two design choices. First, it uses four denoising steps, whereas competing diffusion-based methods typically require tens of steps. Second, it encodes the garment once before denoising and caches its attention features, avoiding repeated processing of garment tokens in subsequent iterations. Other methods instead recompute a dedicated appearance branch or concatenate garment and noisy-image tokens at each step. These choices reduce both sampling and conditioning costs, enabling fast image try-on with the same model used for streaming video generation.
\begin{table}[tb]
    \centering
    \footnotesize
    \renewcommand{\arraystretch}{0.8}
    \setlength{\tabcolsep}{2pt}
    \begin{tabular}{@{}lcccccc@{}}
        \toprule
        & \multicolumn{4}{c}{Paired} & \multicolumn{2}{c}{Unpaired} \\
        \cmidrule(lr){2-5}\cmidrule(l){6-7}
        Method
        & FID$\downarrow$
        & KID$\downarrow$
        & SSIM$\uparrow$
        & LPIPS$\downarrow$
        & FID$\downarrow$
        & KID$\downarrow$ \\
        \midrule
        IDM-VTON      & 6.338 & 1.322 & \underline{0.881} & 0.079 & 9.611 & 1.639 \\
        OOTDiff        & 9.305 & 4.086 & 0.819 & 0.088 & 12.408 & 4.689 \\
        CatVTON         & \underline{6.139} & 0.964 & 0.869 & 0.097 & \textbf{9.143} & \underline{1.267} \\
        UniFit           & 8.799 & \underline{0.702} & \textbf{0.883} & \underline{0.065} & -- & -- \\
        \addlinespace[1pt]
        \rowcolor{blue!5}
        \textbf{Ours}  & \textbf{5.910} & \textbf{0.419} & 0.873 & \textbf{0.054} & \underline{9.212} & \textbf{0.889} \\
        \bottomrule
    \end{tabular}
    \caption{Quantitative comparison of LiveVVT with publicly available image try-on methods on the VITON-HD test set.}
    \label{tab:zalando-test}
\end{table}
\section{Additional Ablation Studies}
\subsection{Analysis of the Hyperparameter $\lambda$}
CoMD combines two complementary supervision signals for recurrent few-step generation. DMD aligns the student distribution with the high-fidelity teacher prior, but provides no direct real-video supervision for the history-conditioned, staggered-noise states that characterize rolling inference. RFM addresses this limitation by sampling windows from real videos and reconstructing the temporal cache, global appearance memory, and staggered noise schedule used during rolling inference. It therefore supervises the student on real-data trajectories under deployment-aligned states. The combined objective,
$\mathcal{L}_{\mathrm{CoMD}}=\mathcal{L}_{\mathrm{DMD}}+\lambda\mathcal{L}_{\mathrm{RFM}}$, balances teacher-prior transfer with real-trajectory alignment, where $\lambda$ controls the contribution of RFM.

Table~\ref{tab:lambda_ablation} evaluates this trade-off on ViViD-SL under paired evaluation. Setting $\lambda=0$ reduces CoMD to DMD-only training. Introducing RFM consistently improves both VFID metrics and SSIM for $\lambda\in\{0.1,0.2,0.4\}$, showing that direct supervision on real rolling trajectories improves video-level fidelity and structural consistency beyond teacher-distribution matching alone. The accompanying increase in LPIPS, however, indicates that RFM should remain complementary to DMD rather than dominate the objective. We adopt $\lambda=0.2$ because it provides the most favorable balance among video fidelity, structural consistency, and perceptual quality. Increasing the weight further produces non-monotonic behavior: $\lambda=0.8$ improves SSIM but weakens both VFID metrics and LPIPS relative to the selected setting. These results validate the collaborative formulation of CoMD and show that a moderate RFM weight best preserves the teacher prior while aligning the student with real-video rolling trajectories.

\begin{table}[tb]
    \centering
    \small
    \renewcommand{\arraystretch}{0.95}
    \setlength{\tabcolsep}{5.0pt}
    \begin{tabular}{@{}crrrr@{}}
        \toprule
        $\lambda$
        & VFID$_I$$\downarrow$
        & VFID$_R$$\downarrow$
        & SSIM$\uparrow$
        & LPIPS$\downarrow$ \\
        \midrule
        0    & 14.515 & 0.205 & 0.826 & 0.094 \\
        0.1  & 12.807 & 0.123 & 0.834 & 0.096 \\
        \rowcolor{blue!5}
        \textbf{0.2} & 13.194 & 0.090 & 0.838 & 0.099 \\
        0.4 & 13.435 & 0.146 & 0.838 & 0.104 \\
        0.8 & 14.446 & 0.146 & 0.843 & 0.104 \\
        \bottomrule
    \end{tabular}
    \caption{Ablation of the RFM weight $\lambda$ in CoMD on ViViD-SL under the paired setting. The highlighted row denotes the default configuration used in the main experiments.}
    \label{tab:lambda_ablation}
\end{table}
\subsection{Cached vs. Repeated Garment Conditioning}
The target garment remains unchanged throughout a try-on sequence, making its appearance features naturally reusable across rolling updates. LiveVVT therefore encodes the garment image once and stores its attention key/value (KV) features in the persistent global appearance memory. To examine whether recomputing the garment condition improves generation, we compare this \emph{Cached} design with \emph{Repeated Encoding}. The latter does not retain garment KV features; instead, it concatenates garment tokens with the noisy tokens of the active window at every rolling denoising iteration and performs joint attention over the extended sequence. Table~\ref{tab:cached_vs_repeated_garment_conditioning} evaluates the resulting quality--efficiency trade-off.

The default Cached design and the Repeated Encoding alternative achieve comparable but metric-dependent generation quality. Cached conditioning performs better on both VFID metrics, while Repeated Encoding yields a small SSIM increase and a lower LPIPS. Recomputing the garment condition therefore provides no consistent advantage across video fidelity, structural similarity, and perceptual quality. Its efficiency cost is more pronounced: throughput decreases from 22.39 to 19.73 FPS, making Cached conditioning $1.13\times$ faster. These results show that persistent garment KV caching removes redundant conditioning computation without compromising overall try-on quality, directly supporting the high-throughput rolling inference targeted by LiveVVT.

\begin{table}[tb]
    \centering
    \small
    \renewcommand{\arraystretch}{0.95}
    \setlength{\tabcolsep}{2.0pt}
    \begin{tabular}{@{}crrrrr@{}}
        \toprule
        Conditioning
        & VFID$_I$$\downarrow$
        & VFID$_R$$\downarrow$
        & SSIM$\uparrow$
        & LPIPS$\downarrow$
        & FPS$\uparrow$ \\
        \midrule
        Cached  & 13.194 & 0.090 & 0.838 & 0.099 & 22.39 \\
        Repeated Encoding & 13.280 & 0.103 & 0.840 & 0.089 & 19.73 \\
        \bottomrule
    \end{tabular}
    \caption{Quality and throughput comparison of garment-conditioning strategies under the paired setting. Cached stores garment KV features in the global appearance memory, whereas Repeated Encoding recomputes garment conditioning at every rolling update.}
    \label{tab:cached_vs_repeated_garment_conditioning}
\end{table}

\section{Additional Visual Comparisons}

\begin{figure}[tb]
\centering
\includegraphics[width=\linewidth]{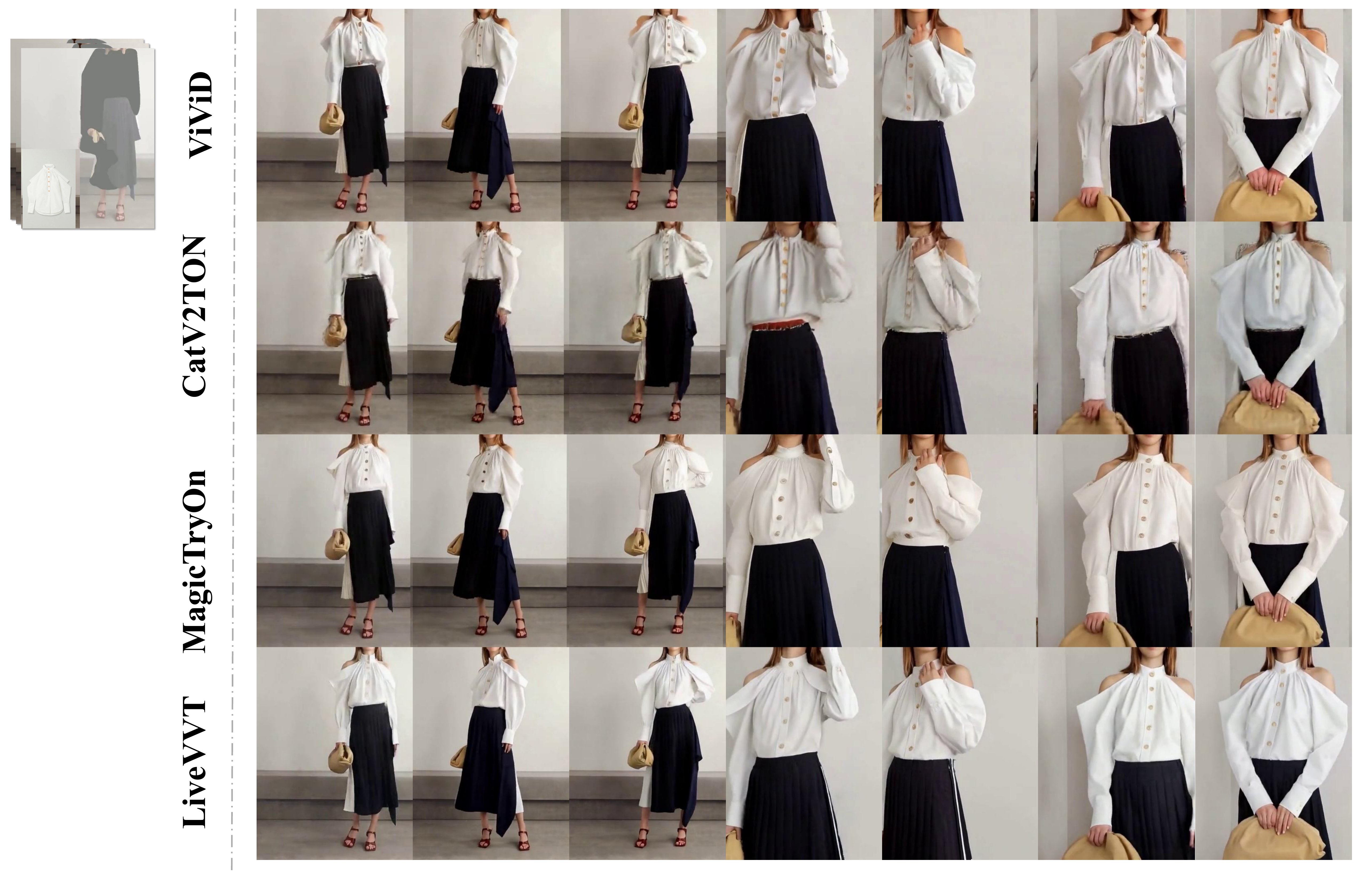}
\caption{Long-sequence comparison for a white blouse under substantial pose and scale changes.}
\label{fig:visual_comparison2}
\end{figure}

\begin{figure}[tb]
\centering
\includegraphics[width=\linewidth]{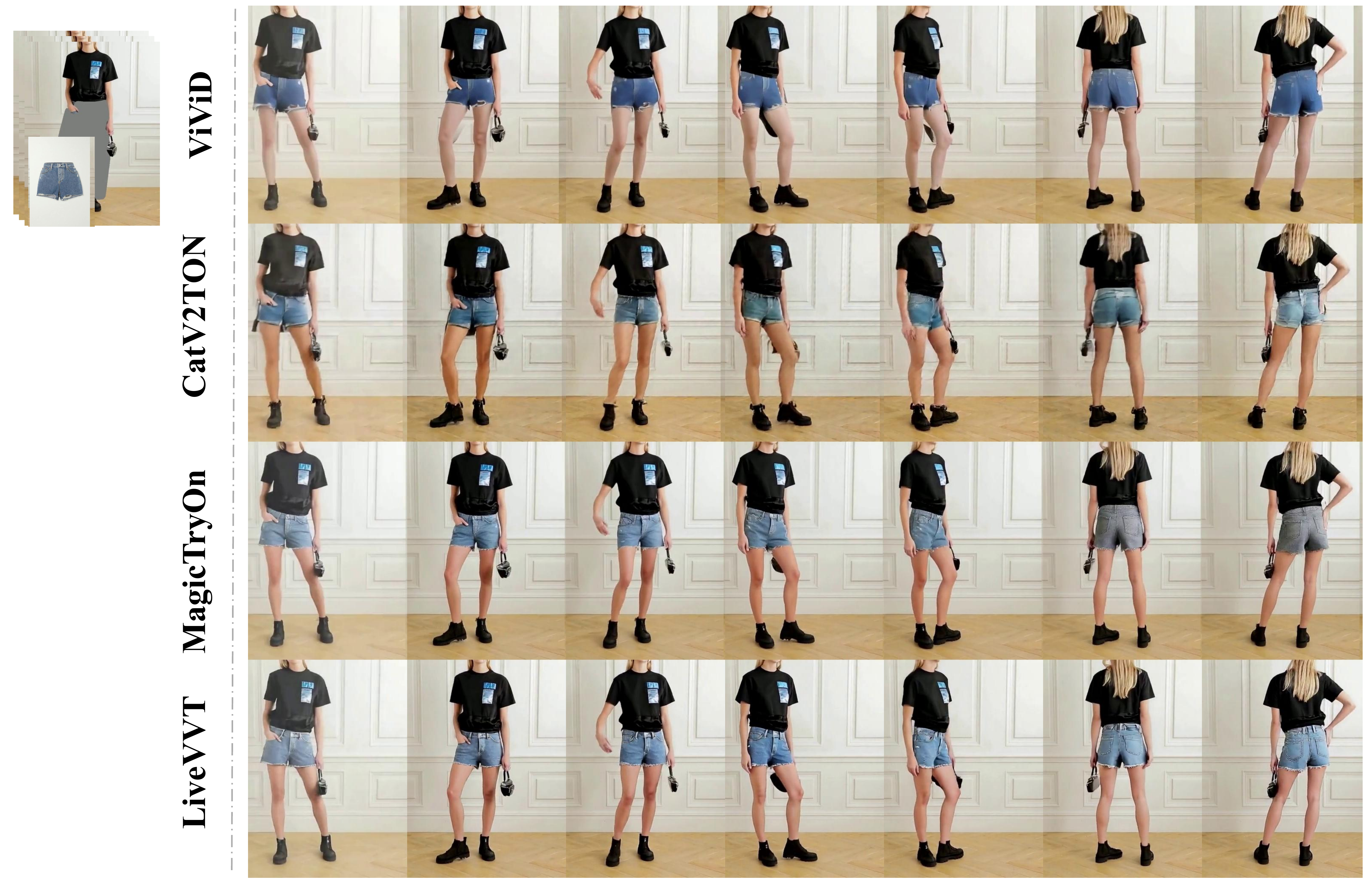}
\caption{Long-sequence comparison for denim shorts across large viewpoint changes.}
\label{fig:visual_comparison3}
\end{figure}

\begin{figure}[tb]
\centering
\includegraphics[width=\linewidth]{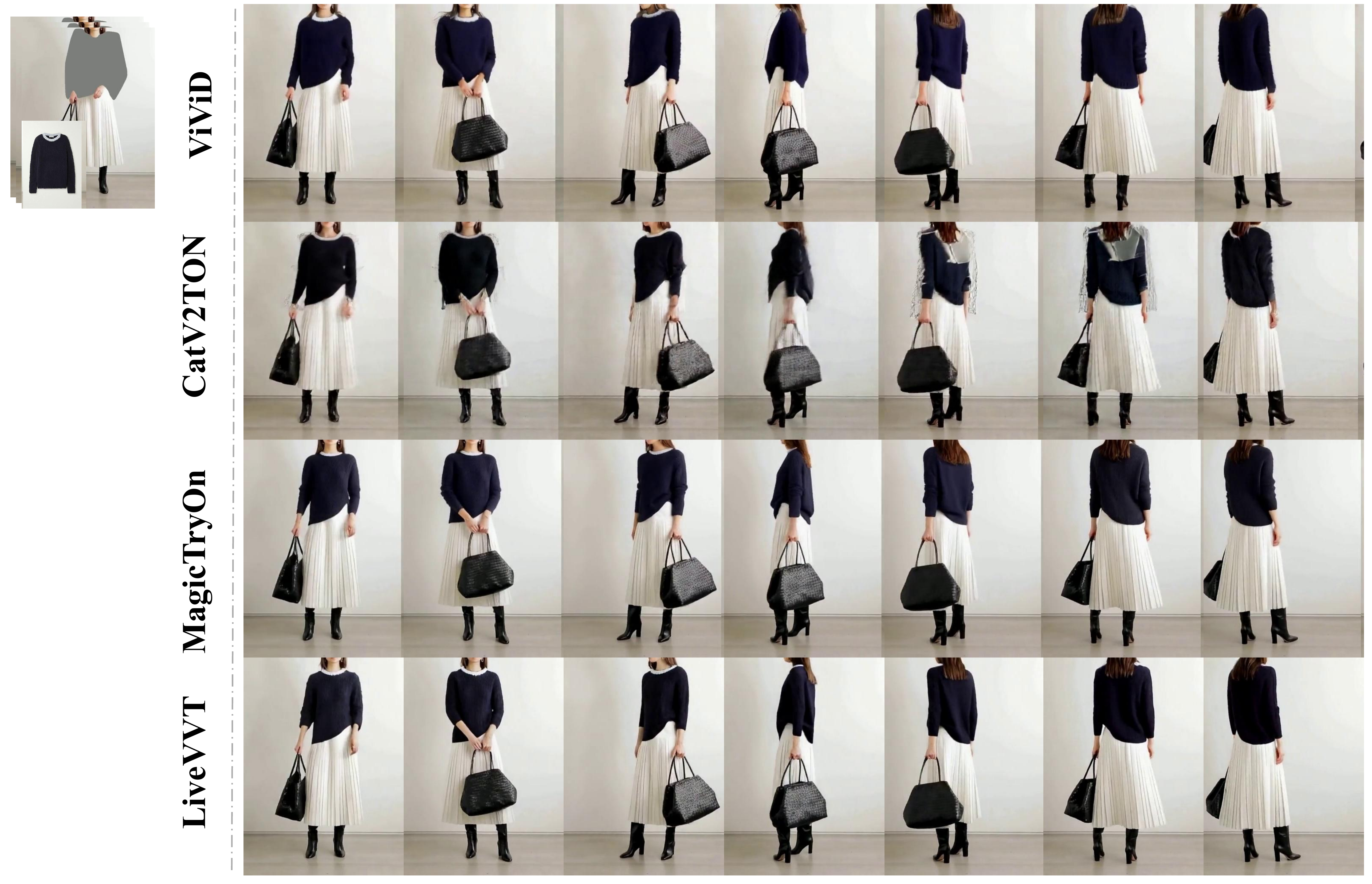}
\caption{Long-sequence comparison for a dark knit top across front and rear views.}
\label{fig:visual_comparison4}
\end{figure}

\begin{figure}[tb]
\centering
\includegraphics[width=\linewidth]{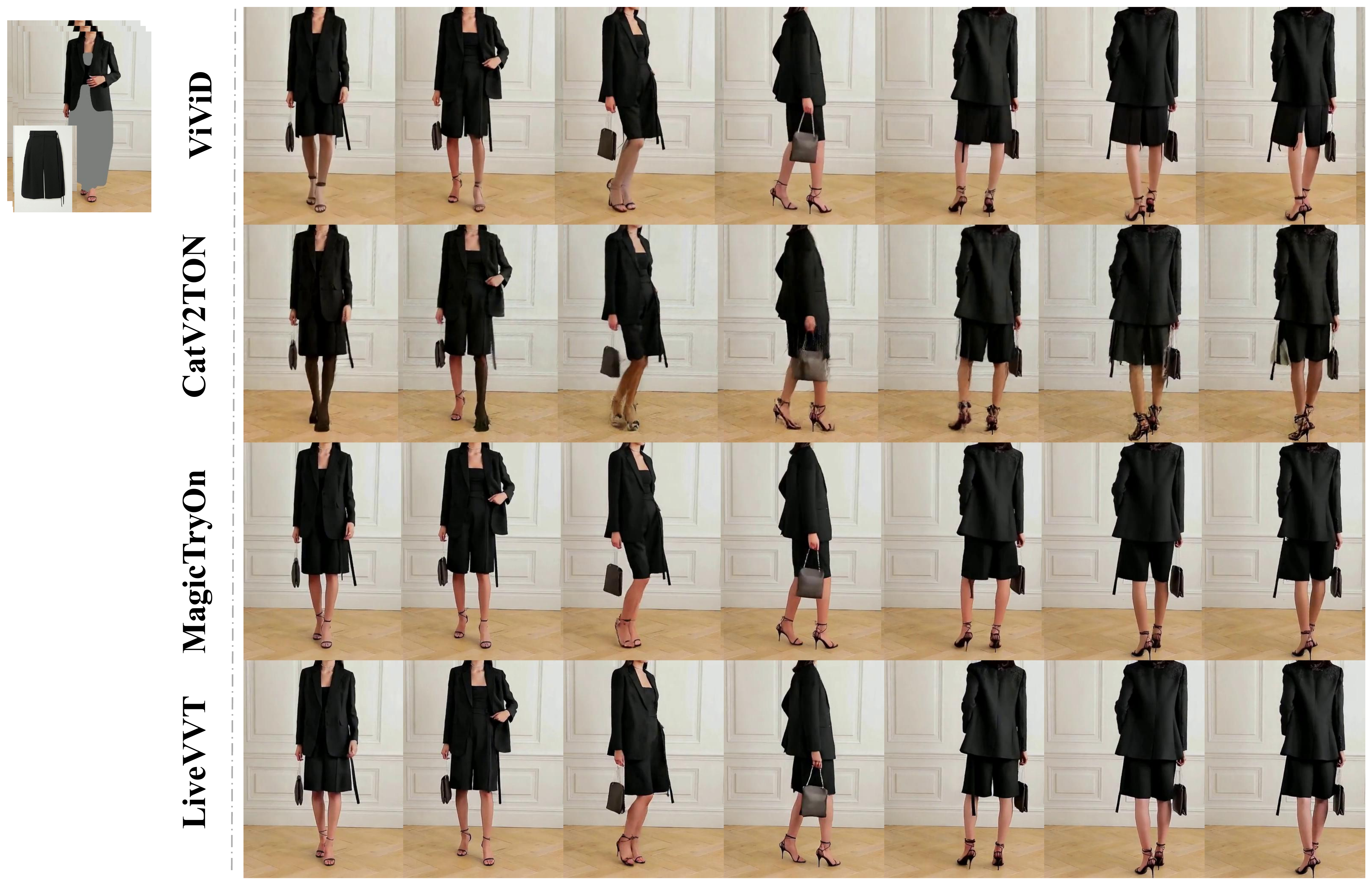}
\caption{Long-sequence comparison for black shorts. MagicTryOn exhibits an appearance shift in later frames, whereas LiveVVT maintains consistent garment color and texture.}
\label{fig:visual_comparison5}
\end{figure}

Figures~\ref{fig:visual_comparison2}--\ref{fig:visual_comparison5} provide additional paired comparisons with publicly available VVT methods, including ViViD, CatV2TON, and MagicTryOn. The examples span multiple garment categories and substantial pose and viewpoint changes, enabling a direct assessment of both local reconstruction fidelity and sequence-level appearance consistency. ViViD and CatV2TON recover the overall garment layout but frequently lose fine texture, exhibit color deviations, or produce boundary artifacts around collars, hems, and occluded regions. These artifacts become more pronounced under large nonrigid deformation, where accurate appearance preservation requires consistent modeling across successive frames.

MagicTryOn produces strong results within individual clips, yet its independently processed windows do not share an explicit sequence-level appearance state. As a result, locally plausible outputs can still undergo texture or color changes at window boundaries. Figure~\ref{fig:visual_comparison5} illustrates this behavior: the generated shorts exhibit a visible color shift in later frames. LiveVVT maintains comparable local detail while preserving a stable garment appearance across the complete sequence. Its rolling window provides local bidirectional interactions for short-range deformation, and the persistent global appearance memory supplies a sequence-level reference that remains fixed across updates. This combination suppresses cross-window drift and yields more coherent long-horizon try-on generation.

\newpage
\section{Long-Sequence Try-On Examples}

\begin{figure*}[tb]
\centering
\includegraphics[width=0.86\linewidth]{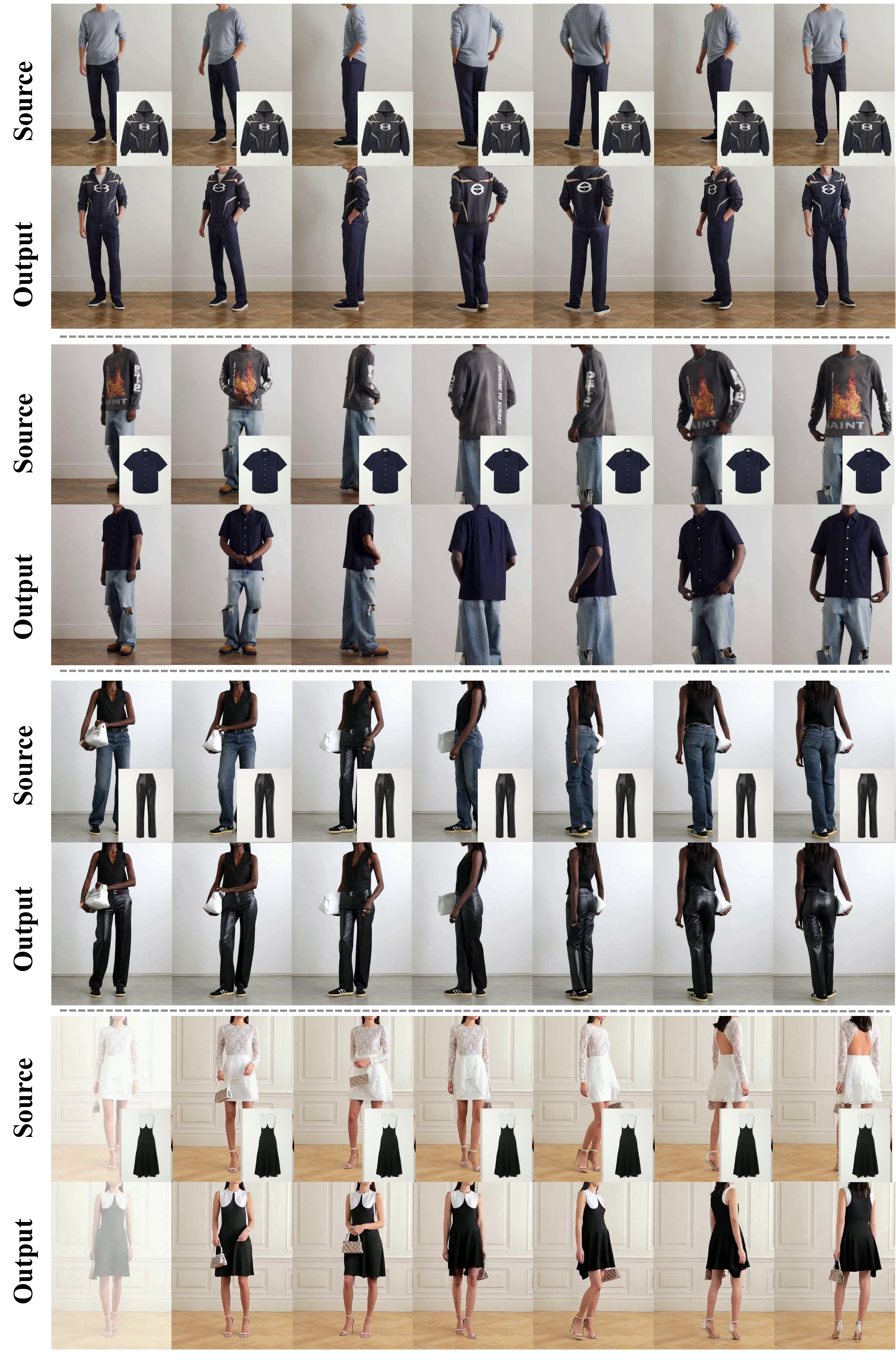}
\caption{Unpaired long-sequence try-on examples across diverse garment categories and viewpoint changes. LiveVVT preserves garment appearance through full-body rotations and extended rolling generation.}
\label{fig:more_examples_1}
\end{figure*}

\begin{figure*}[tb]
\centering
\includegraphics[width=0.86\linewidth]{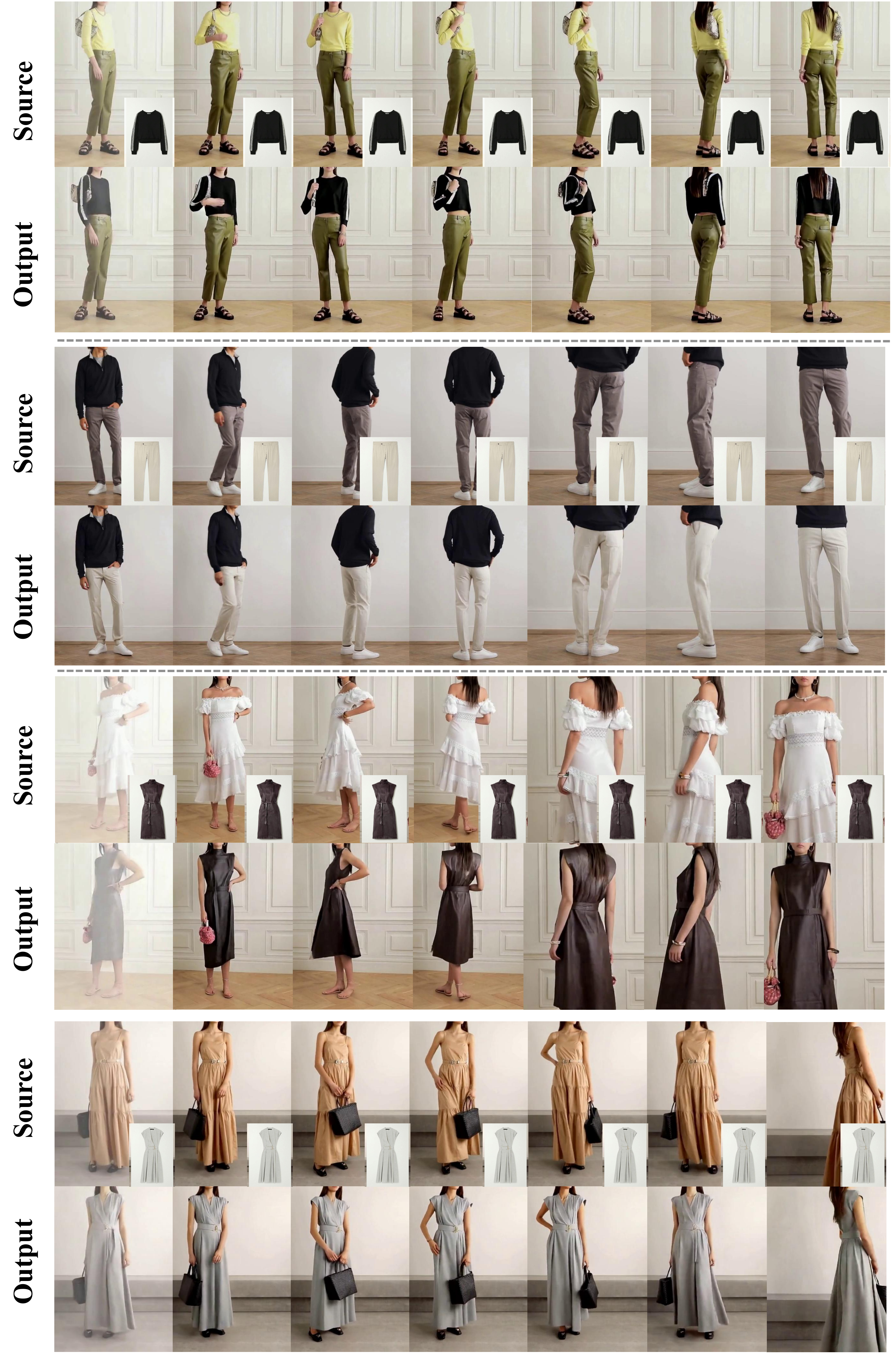}
\caption{Additional unpaired long-sequence examples with substantial viewpoint and camera-scale changes. LiveVVT maintains a coherent dressed appearance after rear views, self-occlusion, and abrupt zooming.}
\label{fig:more_examples_2}
\end{figure*}

Figures~\ref{fig:more_examples_1} and~\ref{fig:more_examples_2} present additional long-sequence results under the unpaired setting, where the target garment is selected independently of the source person's original clothing. The examples cover diverse garment categories, including tops, casual summer outfits, and dresses, together with full-body rotations and substantial changes in subject scale caused by camera motion. These sequences are particularly challenging because the generated garment must remain identifiable when its visible regions change dramatically and must recover consistently after rear views, self-occlusion, or abrupt zooming. LiveVVT maintains coherent garment appearance throughout complete $360^{\circ}$ rotations and preserves a stable dressed appearance under changes in camera scale. This behavior indicates that the persistent global appearance memory provides a sequence-level garment anchor, while rolling-window generation maintains local motion continuity across updates.

\begin{figure*}[tb]
\centering
\includegraphics[width=\linewidth]{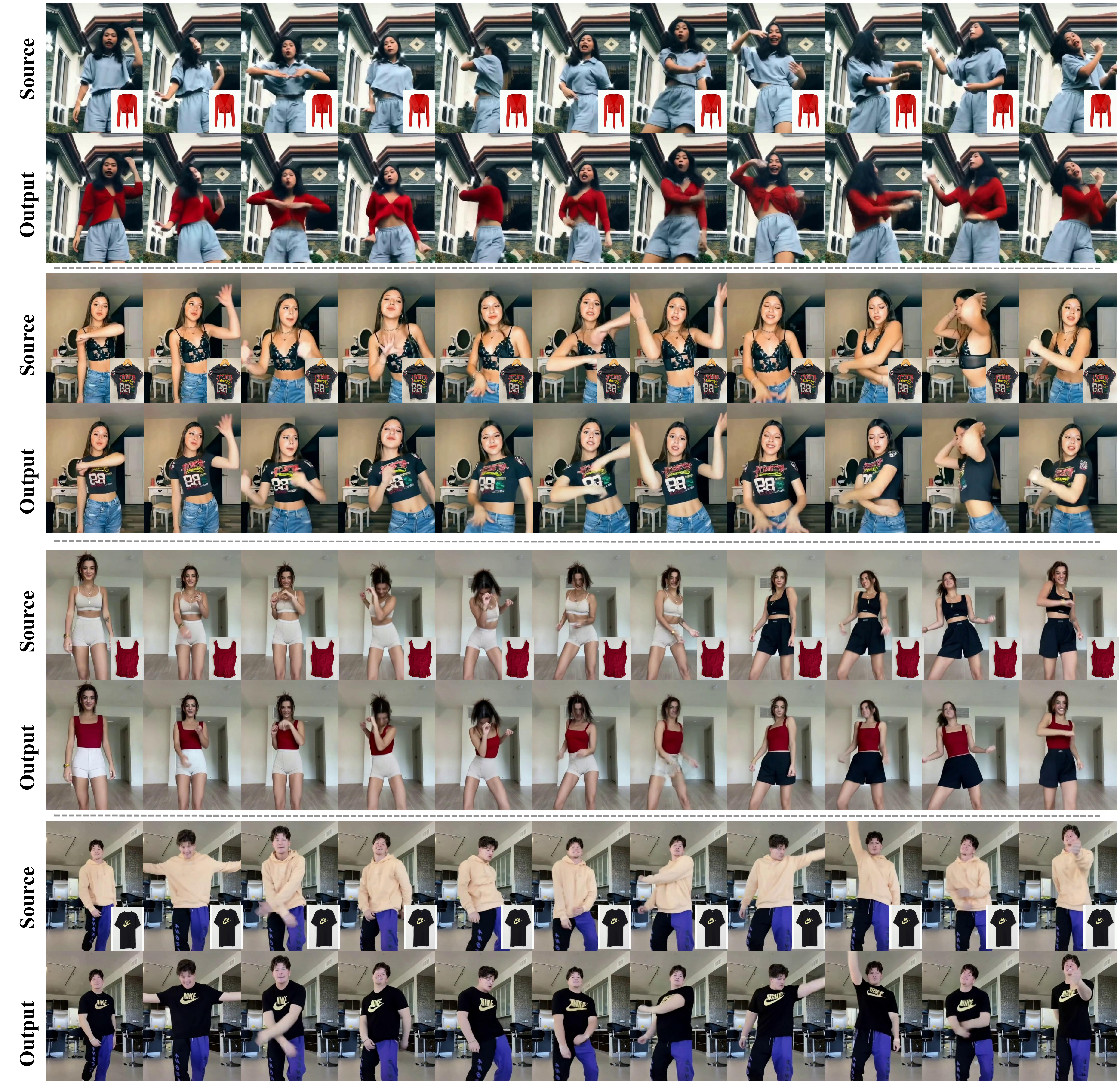}
\caption{Challenging in-the-wild try-on examples with complex backgrounds, rapid motion, and severe self-occlusion. LiveVVT retains robust garment appearance and temporal continuity across these unconstrained sequences.}
\label{fig:more_examples_3}
\end{figure*}

Figure~\ref{fig:more_examples_3} evaluates LiveVVT on in-the-wild videos with complex backgrounds, rapid articulated motion, and severe self-occlusion. Across these challenging cases, the generated results generally preserve the target garment category, appearance, and temporal continuity despite large changes in body configuration. Occasional local artifacts remain around fine garment boundaries and under extreme motion, highlighting directions for further improvement. Nevertheless, the results demonstrate that LiveVVT retains robust long-sequence try-on capability beyond controlled benchmark scenes, supporting its practical use in interactive streaming applications.


\section{Limitations and Discussion}

While LiveVVT substantially improves long-horizon video try-on quality at streaming speed, its design also defines a clear operating regime. The persistent appearance memory is built from the target garment and, for video input, a frontal source-person reference. This anchor is effective for preserving person-specific dressed appearance over extended streams, but the method can become less reliable when the observed appearance departs markedly from the reference because of severe illumination changes, self-occlusion, extreme articulation, or large out-of-plane rotation.

The remaining errors are predominantly localized rather than global: fine garment boundaries, heavily occluded regions, and rapidly moving body parts may still exhibit small distortions or texture inconsistencies. Such failures are only partially reflected by the current evaluation protocol, since SSIM and LPIPS measure frame-level correspondence while VFID measures distributional video quality. A more complete assessment should therefore combine these metrics with explicit temporal-consistency and boundary-accuracy measures, as well as human judgments of garment identity and perceptual stability.

The efficiency gains also involve an explicit context--latency trade-off. Fixed-size rolling windows, cached garment features, and few-step sampling make the per-update cost largely insensitive to sequence length, but they restrict the amount of future context available to each emission and may be suboptimal for abrupt motion or rapid appearance changes. Adaptive windowing or content-aware memory updates could selectively allocate additional computation in such cases while preserving the constant-cost behavior during routine motion. 

\end{document}